\documentclass{article}

\usepackage{tabularx}
\usepackage{arydshln}

\usepackage[utf8]{inputenc}
\usepackage{bm,amsthm,amssymb,amsfonts,mathtools}
\usepackage[margin=1in]{geometry} 
\usepackage{dsfont}
\usepackage{float}
\usepackage{tikz}
\usetikzlibrary{positioning}
\usepackage[framemethod=TikZ]{mdframed}
\usepackage{neuralnetwork}

\usepackage{soul} 
\usepackage{xcolor} 

\colorlet{lighter-red}{red!70}

\usepackage{amsmath,amssymb,amsthm,amsfonts,amscd}
\usepackage{color}
\usepackage[numeric]{amsrefs}
\usepackage{graphicx}
\usepackage{caption}
\usepackage{subcaption}
\usepackage{listings}
\usepackage{makecell}
\usepackage[framemethod=TikZ]{mdframed}

\usepackage{hyperref}

\mdfdefinestyle{TheoremFrame}{%
    linecolor=blue,
    outerlinewidth=1,
    roundcorner=15,
    innertopmargin= \baselineskip,
    innerbottommargin= \baselineskip,
    innerrightmargin=10,
    innerleftmargin=10,
    backgroundcolor=white}
    
\mdfdefinestyle{ProofFrame}{%
    linecolor=red,
    outerlinewidth=1,
    roundcorner=15,
    innertopmargin= \baselineskip,
    innerbottommargin= \baselineskip,
    innerrightmargin=10,
    innerleftmargin=10,
    backgroundcolor=white}    
    
\newlength\tindent
\renewcommand{\indent}{\hspace*{\tindent}}

\newtheorem{numlem}{Lemma}

\theoremstyle{remark}

\newtheorem*{srem}{Remark}

\newtheorem*{nrem}{Remarks}

\title{A Bootstrap Algorithm for Fast Supervised Learning}
\author{Michael A Kouritzin$^1$, Stephen Styles$^2$ and Beatrice-Helen Vritsiou$^1$}
\date{$^1${\footnotesize Department of Mathematical and Statistical Sciences, University of Alberta}\\
	$^2${\footnotesize Statistics Canada, ESMD/DMSE}}

\begin{document}

\maketitle
\begin{abstract}
    Training a neural network (NN) typically relies on some type of curve-following method, such as gradient descent (GD) (and stochastic gradient descent (SGD)), ADADELTA, ADAM or limited memory algorithms. Convergence for these algorithms usually relies on having access to a large quantity of observations in order to achieve a high level of accuracy and, with certain classes of functions, these algorithms could take multiple epochs of data points to catch on. Herein, a different technique with the potential of achieving dramatically better speeds of convergence, especially for shallow networks, is explored: it does not curve-follow but rather relies on `decoupling' hidden layers and on updating their weighted connections through bootstrapping, resampling and linear regression. By utilizing resampled observations, the convergence of this process is empirically shown to be remarkably fast and to require a lower amount of data points: in particular, our experiments show that one needs a fraction of the observations that are required with traditional neural network training methods to approximate various classes of functions. 
\end{abstract}
\section{Introduction}\label{S:intro}

In the last decades, neural networks have proven to be excellent tools in pattern-recognition-type problems, like classification of images, handwriting analysis, stock market prediction, and so on (see e.g.\ \cite{Selv}, \cite{Espana}). Constructing an appropriate neural network for a given application requires first choosing a suitable architecture (including the number of layers and number of nodes for each layer), which can be done based on tried practices as well as theory, and then training it using labelled observations/historical data.  Most commonly, during the training phase, the (weight and bias) parameters of the network are updated using some type of gradient descent (GD).
It is usually desirable to update parameters frequently, after only a portion of the training data, which leads to batch gradient descent (BGD) or stochastic gradient descent (SGD) (in cases where the batches consist of very few data points, or sometimes even a single data point). 
Moreover, it has been observed that adaptive variations of the SGD method help with convergence through boosting the speed of the gradient search and through ensuring that the search does not get stuck in `saddle points' of the parameter space or does not overshoot a point of minimum by too much.  
Adaptive methods such as AdaGrad \cite{DuchiEtAl-2011, McMahan-Streeter-2010}, RMSProp \cite{Tieleman-Hinton-2012}, ADADELTA \cite{Zeiler-2012}, ADAM \cite{Kingma-Ba-2015}, NADAM \cite{Dozat-2016} have collectively shown how SGD can become more responsive to the data it is used on, and are now considered the preferred methods in most (but not necessarily all) applications, being able to handle well many different settings (such as sparse gradients, or a very large number of parameters).

\medskip

The Adaptive Moment Estimation algorithm (ADAM), proposed by Kingma and Ba \cite{Kingma-Ba-2015}, is one of the most commonly used refinements of SGD, combining several ideas of other variations coming before it. Empirically it has been shown to lead to better convergence speeds, and also to perform well regardless of differences in data features between different problems or the choice of neural network (NN) architecture. 
Still, even when using ADAM, many iterations may be required to achieve weight parameter estimates of a desired level of accuracy.
Perhaps, this is due to the fact that ADAM, as well as the other algorithms listed above, are curve-following algorithms, and tracing curves to an extremum (especially without use of higher order derivatives) can be a relatively slow process. On the other hand, where it is known that the function of interest (which in our setting is most commonly the mean squared error (MSE) between the outputs predicted by the NN with current weights and the correct outputs) is at least twice differentiable, then we could potentially use refinements of Newton's method (2nd-order methods) to speed up the convergence. In practice, due to these methods having very clearly increased memory requirements and computational complexity compared to gradient methods, and due to the fact that NNs require a large number of parameters to begin with, quasi-Newton methods, such as the Limited Memory Broyden–Fletcher–Goldfarb–Shanno algorithm (LBFGS) \cite{Liu-Nocedal-1989}, will almost invariably be chosen over them. It should be noted that no optimizer from the abovementioned clearly outperforms all other algorithms in the whole range of even the most standard applications of NNs, so usually an appropriate method is also chosen based on the problem at hand. 

\medskip

In this note, we propose a substantially different technique for training NNs that does not trace curves nor does it even require knowledge of derivatives: this is because it avoids the normal back-propagation step in training NNs. This new method, which we call the Bootstrap Learning algorithm (BLA), appears to work exceptionally well, at least for Single-Hidden-Layer Neural Network (SHLNN),
on which it has been tested.
We believe that it can prove of great use when one has to rely on fewer training data, or when there is need for the training to reach a good level of accuracy within very few iterations. 

The current strong interest in wide NNs \emph{(i.e. few hidden layers, sometimes just one, but potentially many nodes in each of these layers)} is supported by several theoretical results, building on and refining the seminal universal approximation results by Cybenko \cite{Cybenko-1989} and Hornik \cite{Hornik-1991}: in their most basic form these results establish that any well-behaved function with inputs from a compact subset of ${\mathbb R}^d$ and values in ${\mathbb R}^m$ can be approximated to any degree of accuracy by a sufficiently wide SHLNN. 
For this reason and for assimilation purposes, we initially focus on the SHLNN case, but later show how to extend the algorithm to cases where there are more hidden layers.

\smallskip

The proposed algorithm would more closely align to BGD or GD than to SGD in the sense that it uses batches of significant size.
However, instead of gradient calculation/approximation, the  method proposed here
uses a decoupled parameter-updating scheme which relies on a linear approximation method inspired by mathematical results from \cite{Kouritzin-1996a} and \cite{Kouritzin-Sadeghi}. 
To motivate this method, suppose first that $\{x_t:t=1,2,\ldots\}$ and $\{y_t:t=1,2,\ldots\}$ are ${\mathbb R}^d$- and ${\mathbb R}$-valued stochastic processes respectively that satisfy
\begin{equation}\label{process-dependence}
y_t = x_t^Tw + \epsilon_t, \quad t=1,2,\ldots,
\end{equation}
where $ w$ is an unknown $d$-dimensional `weight' vector and $\epsilon_t$ is a stochastic noise sequence. We could think of the vector $x_t$ as being the inputs of a `degenerate' neural network with a single set of connections (that is, no hidden layers), $y_t$ as being the value of one of the output nodes, and $ w$ as being the vector of weights/connections corresponding to this node which we want to accurately determine. In light of these, it would make sense to attempt to minimize the mean squared error
\begin{equation*}\label{eq:mean-squared-error-basic}
 w \ \mapsto\ \lim_{N\rightarrow\infty}\frac{1}{N}\sum_{t=1}^N{\mathbb E}|y_t-x_t^T w|^2
\end{equation*}
(assuming for now that the output sequence $(y_t)_t$ is more or less the correct one). Under some general enough conditions (e.g. assuming that the expectations $A:=\lim\limits_{N\to \infty}\frac{1}{N}\sum\limits_{t=1}^N{\mathbb E}(x_tx_t^T)$ and $b:=\lim\limits_{N\to \infty}\frac{1}{N}\sum\limits_{t=1}^N{\mathbb E}(y_tx_t)$ exist, and that $A$ is positive definite), a unique optimal $ w$ exists and is given by $ w_0:=A^{-1}b$. However, whenever it is difficult to compute/approximate the limiting $A$ and $b$, we can still find a usable estimate for $ w$ by running the following linear approximation algorithm which starts with 
some initial guess $A_0$, $b_0$ and some initial weight $r \ge 0$: set 
\begin{align}
\check{A}_0 &= A_0,\quad  \quad \check{A}_{n+1} = \check{A}_n + \frac{1}{n+1+r}(\underbrace{x_{n+1}x_{n+1}^T}_{A_{n+1}}-\,\check{A}_n), \!\!&n= 0,...,N-1\label{Acheck}
\\
\check{b}_0 &= b_0, \quad \ \ \,\!\!\quad \check{b}_{n+1} = \check{b}_n + \frac{1}{n+1+r}(\underbrace{y_{n+1}x_{n+1}}_{b_{n+1}}-\,\check{b}_n), &n= 0,...,N-1,\label{bcheck}
\end{align}
and then for some small positive constant, or a slowly decreasing sequence of gains $\mu_t$, keep setting 
\begin{align*}\label{eq:weight-update}
 w_{t+1} &=  w_t + \mu_t(\check{b}_N - \check{A}_N w_t)
\end{align*}
until $w_t$ has essentially converged.
In short, we compute the weighted averages
\begin{equation}\label{Combocheck}
\check A_N=\frac{r}{r+N}A_0+\frac{1}{r+N}\sum\limits_{n=1}^NA_n,\quad
\check b_N=\frac{r}{r+N}b_0+\frac{1}{r+N}\sum\limits_{n=1}^Nb_n
\end{equation}
for some large $N$ and then we essentially solve for
$ w_N:=\check A_N^{-1}\check b_N$ in a numerically stable way.
\begin{srem}
To fix/clarify notation that we will be using in the sequel: we will also consider $y$ to be a vector, the vector of all node-values of some layer of the NN; in such cases, we would have \eqref{process-dependence} replaced by $y_t^T=x_t^Tw + \epsilon_t^T$ where $w\in {\mathbb R}^{d\times m}$ is matrix-valued now (and $\epsilon_t$ is a random `noise' vector). Then $b_{n+1}$ in (\ref{bcheck}) would instead be defined by $x_{n+1}y_{n+1}^T$ and $\check{b}_N$ would be ${\mathbb R}^{d\times m}$-valued.
\end{srem}

\medskip
Unless otherwise stated, we will choose $r=N$ for our experiments, which means\begin{equation}\label{Combocheck1}
\check A_N=\frac{1}{2}A_0+\frac{1}{2N}\sum\limits_{n=1}^NA_n,\quad
\check b_N=\frac{1}{2}b_0+\frac{1}{2N}\sum\limits_{n=1}^Nb_n.
\end{equation}
In other words, the initial estimate/`guess', which will be carried over from the previous batch of data used in the algorithm, will be weighted equally as the
estimate produced only from data of the current batch.
This is intended to produce weighted averages that forget earlier estimates at a geometric rate of $\frac12$.

\medskip

Hitherto, we have completely disregarded the crucial facts that to approximate general classes of functions we need to work with NNs with at least one hidden layer and we need the dependence of `inner' nodes on the inputs not to be a linear one as suggested by \eqref{process-dependence}. 
Instead an activation function is also applied (sometimes a different such function per layer) and turns this dependence into a non-linear one. To be able to combine this fact with the linear approximation scheme above, here we propose to `decouple' the NN and apply simultaneous linear approximation processes for every pair of consecutive layers.  In the SHLNN case the first such process would serve to estimate the synaptic weights between the inputs-layer and the hidden layer, and the second one would be for the weights between the hidden layer and the final-outputs-layer. Naturally, this creates a new issue: we have to ensure that the outcomes of the two processes are compatible at all times. This leads to probably the most crucial and novel feature of the proposed method, which is to utilize our training data in two different ways: both for updating, but also for keeping track of the dependence between the two sets of weights produced by the algorithm at each given time (the details of how this is done are briefly presented in the following subsection and are given in full in Section \ref{S:Meth}).

\bigskip

{\bf Notation.} We work with fully-connected NNs which have one hidden layer (in our experiments we choose to have about $m=100$ nodes in this layer, but this can be adjusted as necessary in future applications). For simplicity we will only consider scalar outputs since we can treat multidimensional outputs as separate functions and apply our method on each such function.
We focus on regression problems and use the network architecture in the figure below.

We will count layers from the inputs to the output and use superscripts
and subscripts to distinguish layer and respectively nodes within a layer (or time
depending on context).
$\beta$ will be used for bias (to distinguish from the vectors (matrices) $b$ used in the linear approximation algorithm introduced above), although in most instances we will include it in the vector $w$ of weights as its $w_0$ component (and then we will also set $x_0=-1$ so that the scalar product notation $w^T x$ will include subtracting the bias; it should be clear from the context each time if the bias term(s) is included in the vector (matrix) $w$ or not).
Using vector notation, the picture below, as well as the implied weights and activation functions $\sigma_1,\sigma_2$ (which act component-wise), one finds that
\begin{equation}\label{widenetworkeqn}
\widehat{y}=\sigma_2(w^2 \sigma_1(w^1 x -\beta^1) -\beta^2),
\end{equation}
which by the universal approximation theorem is general enough to approximate any continuous scalar function on a compact subset of $\mathbb R^d$.
Absorbing the bias in $w$ now, we can write
\begin{equation}
\widehat{y}=\sigma_2(w^2 h ),\quad h=\sigma_1(w^1 x ),
\end{equation}
It will also be crucial in the sequel to keep track of the pre-activation values for each layer:
\begin{equation}
{z}^2=w^2 h,\quad z^1=w^1 x .
\end{equation}

\tikzset{%
  every neuron/.style={
    circle,
    draw,
    minimum size=1.5cm
  },
  neuron missing/.style={
    draw=none, 
    scale=4,
    text height=0.333cm,
    execute at begin node=\color{black}$\vdots$
  },
}

\begin{center}

\begin{figure}[H]
    \centering
    \begin{tikzpicture}[x=1.5cm, y=1.5cm, >=stealth]

    \foreach \m/\l [count=\y] in {1,2,missing,3}
    \node [every neuron/.try, neuron \m/.try] (input-\m) at (1.75,2.5-\y*1.25) {};

    \foreach \m [count=\y] in {1,2,missing,3}
    \node [every neuron/.try, neuron \m/.try ] (hiddenA-\m) at (4.25,2.5-\y*1.25) {};

    \foreach \m [count=\y] in {1}
    \node [every neuron/.try, neuron \m/.try ] (output-\m) at (6.75,1.25-\y*1.25) {};

    \foreach \l [count=\i] in {{-1},x_{1},x_{d}}
    \node at (input-\i) {$\l$};

    \foreach \l [count=\i] in {{h_{1}}, h_{2}, h_{m}}
    \node at (hiddenA-\i) {$\l$};
  
    \foreach \l [count=\i] in {1}
    \node at (output-\i) {$\widehat{y}$};

    \foreach \l [count=\i] in {1,2,3}
    \foreach \j in {1,2,3}
    \draw [->] (input-\i) -- (hiddenA-\j);
    \foreach \i in {1,...,3}
    \foreach \j in {1}
    \draw [->] (hiddenA-\i) -- (output-\j);

    \node [align=center, above] at (1.75,2) {Input};
    \node [align=center, above] at (4.25,2) {Hidden Layer};
    \node [align=center, above] at (6.75,2) {Output};
    
    \end{tikzpicture}
    \caption{Neural network architecture}
    \label{fig:NN}
\end{figure}
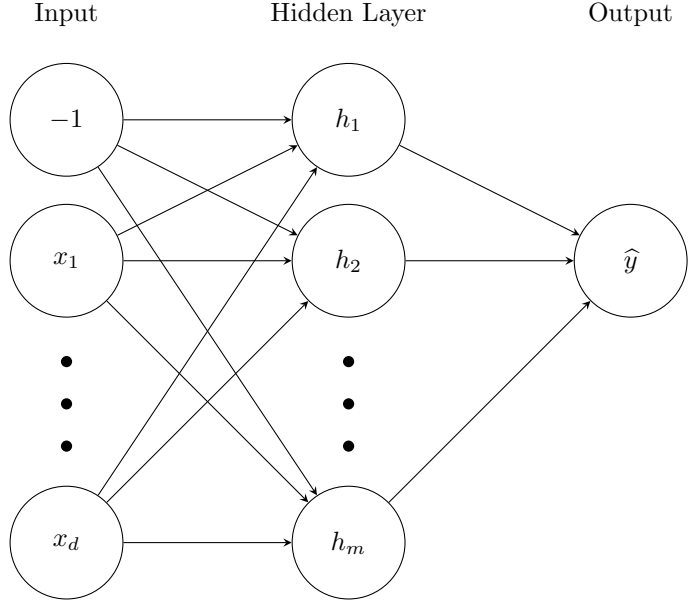

\end{center}

\subsection{Bootstrap Particles}

Pretend for the moment that we have access to (accurately found) hidden values $h$.  (Of course, this is not assumed in our work.)  
Then, the internal variables $h_n$ (corresponding to a given input $x_n$ of a function $f$ that we want to approximate, with $f(x_n)=y_n$ being the actual output for $x_n$) can be viewed as the input of layer 2 and their pre-activation values $z_n^1$ (which will equal $\sigma_1^{-1}(h_n)$ when the activation function $\sigma_1$
happens to be invertible) become the outputs for layer 1.
Note that $x_n$ (the actual input to the NN) and $z_n^1$ satisfy a linear relation, which we want to optimize using the linear approximation algorithm introduced above. Similarly, $h_n$ and $z_n^2$ satisfy another linear relation (the latter values equalling $\sigma_2^{-1}(y_n)$ if $\sigma_2^{-1}$ exists; in our experiments we primarily set $\sigma_2$ equal to the identity, which is also a standard choice in linear regression). We have thus decoupled the SHLNN into two `degenerate' NNs where our earlier linear approximation scheme applies:\medskip \\
\noindent Layer 1 to 2:
$\ \displaystyle \begin{array}{rcl}
    A^1_n &= &x_n {x_n}^T\\
    b^1_n &=  &{x_n}(z^1_n )^T
\end{array}$\medskip\\
Layer 2 to 3:
$\ \displaystyle \begin{array}{rcl}
    A^2_n &= &h_n {h_n}^T\\
    b^2_n &=   &{h_n}(\sigma_2^{-1}(y_n))^T
\end{array}$.\medskip\\
\noindent However, this presupposes that we have access to at least a fairly decent approximation of $h_n$ and $z_n^1$, which in reality we cannot guarantee.

We deal with this issue by creating {\em bootstrap particles}. Given the current value of the weights (and biases), the bootstrap particles are created by running each input (coming from a batch of the data of size $N$) through the network and by keeping track of the internal values and the final output: let $\{(x_n,z^1_n,h_n,z^2_n,\hat{y}_n)\}_{n=1}^N$ be the 5-tuples/particles we get in this way (note that in general $\hat{y}_n \neq y_n$, the actual output of the function $f$ for input $x_n$).
To keep notation clear, we relabel the particles $\left\{\bigl(x_{n},z^1_n,h_n,z^2_n,\hat y_{n}\bigr)\right\}_{n=1}^{N}$ as  $\bigl\{\bigl(\hat x_{k},\hat z^1_k,\hat h_{k},\hat z^2_k,\hat y_{k}\bigr)\bigr\}_{k=1}^{N}$: these will help us pick suitable internal values $\hat{h}_{s_n}$ and $\hat{z}_{s_n}^1$, which we can combine with the real data $\left\{\bigl(x_{n},y_{n}\bigr)\right\}_{n=1}^{N}$ to then apply the linear approximation scheme. 
In particular, for each $(x_n,{y}_n)$, we consider the $\delta$ \emph{closest} particles (i.e. the $\delta$ \emph{smallest} distances among $\ell_{n\rightarrow k}=\|(x_n,{y}_n)-(\hat x_k,\hat{y}_k)\|$, $k\in\{1,...,N\}$) and use a (scoring) function $s$ that is large when $(x_n,y_n)$ is close to $(\hat x_{k},\hat y_{k})$. 
With probability proportional to this scoring function,
some bootstrap particle $\bigl(\hat x_{k_j},\hat z^1_{k_j},\hat h_{k_j},\hat z^2_{k_j},\hat y_{k_j}\bigr)$, $1\le j\le \delta$, is accepted for the real data pair $(x_n,y_n)$, and this allows us to form the `mixed' particles 
\begin{equation*}
\left\{\left(x_n,\,\hat z^1_{k_{j_n}},\,\hat h_{k_{j_n}},\ z^2_n = \sigma_2^{-1}(y_n),\, y_n\right)\right\}_{n=1}^N.
\end{equation*}
\begin{nrem}
1. For notational simplicity, from now on we will write $\hat{h}_n$ instead of $\hat h_{k_{j_n}}$ (and similarly $\hat{z}_n^1$ instead of $\hat z^1_{k_{j_n}}$); in particular the notation $\hat{h}$ and $\hat{z}$ will indicate that these node-values are coming from the bootstrap particle drawn for the real data pair $(x_n,y_n)$, and are not, in general, equal to the internal values we would get if we fed $x_n$ into the NN (with current weights).
    \medskip\\
    2. As already mentioned, in our experiments we set $\sigma_2$ to be the identity function, so we can safely set $z^2_n = \sigma_2^{-1}(y_n)$. If on the other hand in some application we had to work with a non-invertible activation function for the last layer, we could instead pick $z_n^2$ as follows: for each $(x_n,y_n)$ we could draw a second bootstrap particle (particle $k_{u_n}$), in the same way as above but independently of the first particle $k_{j_n}$, and we could set $z^2_n = \hat{z}_n^2 = \hat z^2_{k_{u_n}}$.
\end{nrem}

\medskip

\noindent We can now safely apply the linear approximation scheme described earlier for each pair of consecutive layers of our SHLNN:
\\
\medskip

\noindent Layer 1 to 2:
$\ \displaystyle \begin{array}{rcl}
    A^1_n &=&x_n x_n^T\\
    b^1_n &=&x_n (\hat z^1_n)^T
\end{array}\qquad\leadsto \quad w^1=\left[A^1_N\right]^{-1}b^1_N$\medskip\\
Layer 2 to 3:
$\ \displaystyle \begin{array}{rcl}
    A^2_n &= &\hat h_n {\hat h_n}^T\\
    b^2_n &=  &{\hat h_n}(z^2_n)^T
\end{array}\qquad\leadsto \quad w^2=\left[A^2_N\right]^{-1}b^2_N$.
\bigskip\\
\noindent \emph{Remark on the methodology.} We primarily use the squared $l_2$ distance: $\ell_{n\rightarrow k}=\left|x_n-\hat x_{k}\right|^2+\left|y_n-\hat y_{k}\right|^2$, and the scoring function $s(l)=e^{-l^2}$. In our experiments we started by setting $\delta=\delta_m = 40$, and gradually decreased
this all the way down to $8$, as the number $m$ of batches of data we had processed increased (once $\delta$ reached 8, we kept it static). 

In principle, we expect that allowing $\delta$ to decrease as the training progresses (even all the way down to 1) can improve performance, and it should also help with establishing theoretical results. However, we did not explore in this work which decreasing function(s) $m\rightarrow\delta_m$ would be best to choose.

\medskip

The rest of the paper is organized as follows: in Section \ref{S:Meth} we give precise details of the method (and some theoretical considerations), while in Sections \ref{S:Reg} and \ref{S:Class} we describe our experiments and present the comparison with other methods (the methods we are comparing to are gradient descent (GD),  ADAM and the Limited Memory Broyden–Fletcher–Goldfarb–Shanno algorithm (LBFGS)).

\bigskip

\noindent {\bf Acknowledgement.} This work builds on some ideas put forth while the second-named author was a Master's student, and later Research Associate, at the University of Alberta.

\section{Method} \label{S:Meth}

As already mentioned, in this paper we largely focus on networks that only contain one hidden layer.  For this architecture we need to choose two activation functions $\sigma_1$ and $\sigma_2$, where $\sigma_1$ is the activation function between the input layer and the hidden layer, and $\sigma_2$ is the activation function between the hidden layer and the output layer. Typically, for regression one chooses $\sigma_2$ to be the identity function, which is consistent with the universal approximation theorem. For the purposes of this paper we mostly use the hyperbolic tangent for our first activation function $\sigma_1$.  However, ReLU and Leaky ReLU are
also used where more appropriate.

\medskip

We consider a single set of training data with $\mathcal N$ total input-output observations, 
which we can (and intend to) reuse multiple times, with each time through the data being called
an epoch. Each epoch $e$ will be split into one or more (disjoint) batches, labelled $M_l(e),...,M_h(e)$ from
lowest to highest using contiguous numbers, so that $M_l(1)=1$, and $M_l(e+1)=M_h(e)+1$ for $e\ge 1$.
The size of batch $m$ will be denoted by $N_m\le \mathcal N$, or more simply $N$ if there is no ambiguity, and it can vary from epoch to epoch (and even for different batches within the same epoch).

\medskip

Below are the steps the method follows for each batch of data.
\begin{description}
\item[Weight Initialization:]
To initialize the algorithm, we first generate random weights for the network. 
This can be done in many ways but we will just use a standard approach by generating independent weights from a normal distribution centered at zero with small variance (in our case we use a variance of $0.5$).  The key consideration is not to set the weights to zero.
\item[Bootstrap Particle Proposals:]
At the start of every batch, each input in the batch is fed forward into the network, and we keep track of the inputs, hidden node-values, and estimated output. 
We denote these values by $\{(\hat x_i,\hat z^1_i,\hat h_i,\hat z^2_i,\hat y_i)\}_{i=1}^N$.
Of course, they depend on what the current weights (and biases) of the network are.
\item[Scoring:] One data point at a time (from within the current batch), we compare the pair of its input value and true output value $(x,y)$ to the corresponding components of all $(\hat x_i,\hat z^1_i,\hat h_i,\hat z^2_i,\hat y_i)$, $1\le i\le N$, using a norm function. (For this paper, we use the (squared) $\ell_2$ norm for single-input networks and the $\ell_\infty$ norm for multiple inputs.) 
We input this norm distance into a scoring function $s$ that assigns a higher score to units with smaller norm. 
\item[Probability Distribution Creation:]
For each real data point, we select the Bootstrap Particle Proposals with the $\delta$ highest scores $\{(\hat x_i,\hat z^1_i,\hat h_i,\hat z^2_i,\hat y_i)\}_{i=1}^\delta$ and create a probability distribution using the scores as relative weights. 
Note that each real datum is assigned its own probability distribution.
\item[Bootstrap Particle Acceptance:]
We now sample from each probability distribution corresponding to an actual input-output pair $(x_n,y_n)$, $1\le n\le N$, and record {\bf the internal values} of the sampled bootstrap particle, which are combined together with the values $x_n$ and $y_n$, to form the new proxy data point for $(x_n,y_n)$ in the regression computations below. More precisely, we replace $(x_n,y_n)$ by the `mixed' particle $$(x_n, \hat{z}_n^1, \hat{h}_n, \sigma_2^{-1}(y_n), y_n),$$ where $\hat{z}_n^1$ and $\hat{h}_n$ are the corresponding internal values of the sampled bootstrap particle. (Recall also that $\sigma_2=$ identity function here, so $\sigma_2^{-1}(y_n) = y_n$.) Because of the norming and scoring previously done, we have created a proxy data point with internal variables consistent with the actual input-output datum $(x_n,y_n)$ with high likelihood.
\item[Regression Weight Update:]
Now that we have the proxy values $\{(x_n,\hat z^1_n,\hat h_n,z^2_n,y_n)\}_{n=1}^N$, we can update the weights and biases using the batch method from section \ref{S:intro}. 
That is, we update both $\check{A}_n$ and $\check{b}_n$ using the proxy data points for the current batch, and then solve for the weights (see the equations below).
\end{description}

\newpage

\noindent
\hrulefill \\
\vspace{-0.1cm}

\textbf{Algorithm} Batchmode Bootstrap Learning  \\
\vspace{0.05cm}
\noindent\hrulefill \\
\vspace{0.1cm}
\textbf{For} epoch $e=1$ to $E$ \qquad    /* $w^1_1$ and $w^2_1$ were already randomly set */\\[0.25ex]
\indent \textbf{For} batch $m=M_l(e)$ to $M_u(e)$ \\[0.25ex]
\indent \indent Let $N=N_m$ be size of batch $m$.\\[0.25ex]
\indent \indent \textbf{For} $n=1$ to $N_m$ \\[0.25ex]
\indent \indent \  $\hat z^1_n=w^1(m) \hat x_n -\beta^1$,\ $\hat h_n = \sigma_1(\hat z^1_n)$\ \, /* Feed $x$ forward to determine Bootstrap proposals */ \\[0.25ex]
\indent \indent \  $\hat z^2_n=w^2(m) \hat h_n - \beta^2$,\ $\widehat{y}_n = \sigma_2(\hat z^2_n)$\\[0.25ex]
\indent \indent \textbf{For} $n=1$ to $N_m$ \ \qquad /* Find Bootstrap target particles */ \\[0.25ex] 
\indent \indent \indent\textbf{For} $i=1$ to $N_m$ do \qquad  /* Find distances from each datum */ \\[0.25ex]
\indent \indent \indent \indent $\ell_{n\rightarrow i}=||(x_n,y_n)-(x_i,\widehat{y}_i)||$\\[0.25ex]
\indent \indent \indent Sample and create $(x_n,{\bf \hat z^1_n},{\bf \hat h_n},z^2_n,y_n)$ with $\{s(\ell_{n\rightarrow i})\}_{i=1}^{N_m}$ as weights for $\{(\hat x_i,\hat z^1_i,\hat h_i,\hat z^2_i,\hat y_i)\}_{i=1}^{N_m}$ \\[0.25ex]
\indent \indent \textbf{If} $e$ $=$ $1$ then $r=0$, else $r=N_m$ and set\\[0.25ex]
\indent \indent $\check{A}^1_N(m)=\frac{r}{r+N_m}\check{A}^1_N(m-1)+\frac{1}{r+N_m}\sum\limits_{n=1}^{N_m} x_n x_n^T$,\\[0.25ex]
\indent \indent  $\check{b}^1_N(m)=\frac{r}{r+N_m}\check{b}^1_N(m-1)+\frac{1}{r+N_m}\sum\limits_{n=1}^{N_m} x_n (\hat z^1_n)^T$,\\
\indent \indent $\check{A}^2_N(m)=\frac{r}{r+N_m}\check{A}^2_N(m-1)+\frac{1}{r+N_m}\sum\limits_{n=1}^{N_m}\hat h_n\hat h_n^T$,\\
\indent \indent  {$\check{b}^2_N(m)=\frac{r}{r+N_m}\check{b}^2_N(m-1)+\frac{1}{r+N_m}\sum\limits_{n=1}^{N_m} \hat h_n(z^2_n)^T$}\\[0.25ex]
\indent \indent Set $\mu_t^1$ = $2/[\max\{\lambda(\check{A}^1_N(m))\} + \min\{\lambda(\check{A}^1_N(m))\}  ]$\\[0.35ex]
\indent \indent Set $\mu_t^2$ = $2/[\max\{\lambda(\check{A}^2_N(m))\} + \min\{\lambda(\check{A}^2_N(m))\}  ]$\\[0.35ex]
\indent \indent /* Get weights and biases for new epoch/batch within epoch */\\[0.35ex]
\indent \indent Set $t=0$ and 
$w^i(m)=\left\{\begin{matrix}0&m=1\\w^i(m-1)&m>1
\end{matrix}\right.$ for $i=1,2$\\
\indent \indent \textbf{While} still converging do \\[0.35ex]
\indent \indent \indent $w^1(m) = w^1(m) + \mu_t^1(\check{b}^1_N(m) - \check{A}^1_N(m)w^1(m) )$\\[0.35ex]
\indent \indent \indent $w^2(m) = w^2(m) + \mu_t^2(\check{b}^2_N(m) - \check{A}^2_N(m)w^2(m) )$\\[0.35ex]
\indent Shuffle the data for the next epoch.\\
\vspace{0.15cm}
\noindent\hrulefill
\vspace{0.2cm}

\begin{nrem}
1. Most quantities implicitly depend upon the batch number $m$, for example $\hat h_n=\hat h_n(m)$ and $\delta=\delta_m$, or the iteration number $t$, for example $w^1(m) =w^1_t(m)$.
However, most of these dependencies are suppressed so as not to overcomplicate the notation in the algorithm.\smallskip\\
2. It should be noted (see also the following lemma) that, for $w^i(m), i=1, 2$ to be guaranteed to converge (with the limit being equal to $[\check{A}^i_N(m)]^{-1}\check{b}^i_N(m)$), 
we need to ensure the eigenvalues of
$I-\mu_t^2 \check{A}^i_N(m)$ are inside the unit ball.
This amounts to the two conditions: 
\begin{align*}
    \mu_t^i < \frac{2}{\max\{\lambda(\check{A}_N^i(m))\}} \ \text{ and }\ \mu_t^i\min\{\lambda(\check{A}_N^i(m))\}>0
\end{align*}
but the second condition is always true if $\mu_t^i>0$ and $\check{A}_N^i(m)$
is nonsingular (with the latter holding with high probability as long as each batch of data has sufficiently large size).
Still, to center the extreme eigenvalues in the unit disk (and having experimentally observed higher speed of convergence with the choice below), in practice we take
$$\mu_t^i = 1.95/[\max\{\lambda(\check{A}^i_N(m))\} + \min\{\lambda(\check{A}^i_N(m))\}].
$$
Note also that all the linear processes we ran in our experiments appear to converge relatively quickly 
even in cases where we observed the matrices $\check{A}^i_N(m)$ to be ill-conditioned. This may be worth exploring further from a theoretical standpoint too.

Currently, we compute $\lambda(\check{A}^i_N)$ efficiently using default R packages given that our matrices are only of size $100\times100$ at most (with this size attained by the matrices constructed from the hidden nodes), but more sophisticated methods should be explored for larger datasets (i.e. more hidden nodes). 
Alternatively, one could just set $\mu_t^1(t)\searrow 0$ slowly at the cost of slightly slower convergence (see the following lemma as well).  For example, we could take $\mu_t^i(t)=\frac1{\ln(1+t)}$.
\end{nrem}

We now provide some theoretical justification for the ``while still converging'' part in the final step of each epoch.

\begin{numlem}\label{Lem0} Suppose that all the eigenvalues of $[I-\mu_t^1\check{A}^1_N(m)]$
and $[I-\mu_t^2\check{A}^2_N(m)]$ are (eventually) strictly inside the unit ball and that $\mu_t^1,\mu_t^2 > 0$ satisfy $\sum\limits_{t=1}^\infty \mu_t^i=\infty$ for $i=1,2$. Then, the lines\\
\indent \indent \textbf{While} still converging do \\[0.35ex]
\indent \indent \indent $w^1_{t+1}(m) = w^1_t(m) + \mu_t^1(\check{b}^1_N(m) - \check{A}^1_N(m)w^1_t(m) )$\\[0.35ex]
\indent \indent \indent $w^2_{t+1}(m) = w^2_t(m) + \mu_t^2(\check{b}^2_N(m) - \check{A}^2_N(m)w^2_t(m) )$\\[0.35ex]
really solve $w^1(m)=\left[ \check{A}^1_{N}(m)\right]^{-1}\check{b}^1_{N}(m) $ and $w^2(m)=\left[ \check{A}^2_{N}(m)\right]^{-1}\check{b}^2_{N}(m) $ as $t\rightarrow\infty$.
\end{numlem}
\proof
Consider only the first equation as the proof for the second one is identical.  One has
\begin{equation}
    w^1_{t+1}(m) = w^1_{t}(m) + \mu_t^1(\check{b}^1_N(m) - \check{A}^1_N(m)w^1_{t}(m) ).
\end{equation}
Hence, if we subtract $\left[ \check{A}^1_{N}(m)\right]^{-1}\check{b}^1_{N}(m)$
from both sides and simplify, we get
\begin{align}
     w^1_{t+1}(m)-\left[ \check{A}^1_{N}(m)\right]^{-1}\check{b}^1_{N}(m) &= w^1_{t}(m) -\left[ \check{A}^1_{N}(m)\right]^{-1}\check{b}^1_{N}(m)\\\nonumber
     &\qquad + \mu_t^1\check{A}^1_N(m)(\left[ \check{A}^1_{N}(m)\right]^{-1}\check{b}^1_N(m) - w^1_{t}(m) )\\\nonumber
     &=(I-\mu_t^1\check{A}^1_N(m))\left(w^1_{t}(m) -\left[ \check{A}^1_{N}(m)\right]^{-1}\check{b}^1_{N}(m)\right).
\end{align}
Now, using recursion as well as standard methods (see \cite{Kouritzin-1994}, \cite{Kouritzin-1996a}, \cite{Kouritzin-Sadeghi}), we can check that   
\begin{align*}
     \|w^1_{t+1}(m)-\left[ \check{A}^1_{N}(m)\right]^{-1}\check{b}^1_{N}(m) \|
     &=\Bigl\|\prod_{k=1}^t(I-\mu_k^1\check{A}^1_N(m))\left(w^1_{1}(m) -\left[ \check{A}^1_{N}(m)\right]^{-1}\check{b}^1_{N}(m)\right)\Bigr\|\\\nonumber
     &\le\prod_{k=1}^t\|I-\mu_k^1\check{A}^1_N(m)\|\,\cdot \left\|w^1_{1}(m) -\left[ \check{A}^1_{N}(m)\right]^{-1}\check{b}^1_{N}(m)\right\|\\\nonumber
     &\le C\prod_{k=1}^t\exp\left(-\mu_k^1\lambda_m\vee (\mu_k^1\lambda_M-2)\right)\\\nonumber
     &\quad\longrightarrow\  0 \text{ as }t\rightarrow\infty.
\end{align*}
Here $C>0$ is some absolute constant and $\lambda_m,\lambda_M$ are the minimum, maximum 
eigenvalues of $\check{A}^1_N(m)$ respectively.
\endproof

\begin{nrem}
   1. The above proof also establishes that the convergence is geometrically quick if $\mu_t^1,\mu_t^2$ are kept constant (in $t$).
   \smallskip\\
2. Because of the lemma, in practice we can replace the line ``while still converging'' of the pseudocode by some condition on the distance between consecutive updates of $w^i(m)$ becoming smaller than some preselected tolerance. For transparency, in our experiments we just ran each linear process corresponding to batch $m$ for a fixed number of iterations ($100,000$ iterations). 
\end{nrem}


\subsection*{Training a NN with two hidden layers}

There is only one crucial change we should make with the addition of another hidden layer.
There are now two types of hidden variables of the form $h^i$: $h^1$ which is closer to the input layer and
$h^2$ which is closer to the output layer. There are also three (unknown) pre-activation variable types $z^1, z^2, z^3$, which we can also write as 
$\sigma_1^{-1}(h^1),\sigma_2^{-1}(h^2),\sigma_3^{-1}(\hat y)$ respectively, when the
activation functions $\sigma_1,\sigma_2,\sigma_3$ are invertible. For the algorithm we should set $z_n^3 = \sigma_3^{-1}(y_n)$, assuming we pick an invertible $\sigma_3$, where $y_n$ is the actual output for input $x_n$. For the other internal variables, we now sample independently two bootstrap particles 
\begin{equation*}
\left(\hat x_{k_{j_n}}, \hat z^1_{k_{j_n}}, \hat h^1_{k_{j_n}}, \hat z^2_{k_{j_n}}, \hat h^2_{k_{j_n}}, \hat z^3_{k_{j_n}}, \hat y_{k_{j_n}}\right) \quad \hbox{and}\quad  \left(\hat x_{k_{u_n}}, \hat z^1_{k_{u_n}}, \hat h^1_{k_{u_n}}, \hat z^2_{k_{u_n}}, \hat h^2_{k_{u_n}}, \hat z^3_{k_{u_n}}, \hat y_{k_{u_n}}\right),
\end{equation*} 
and create the `mixed' particle
\begin{equation*}
\left(x_n,\ \hat z^1_{k_{j_n}},\hat h^1_{k_{j_n}},\ \hat z^2_{k_{u_n}}, \hat h^2_{k_{u_n}},\ z^3_n,\, y_n\right),
\end{equation*}
which we use in the learning of the first, second and third sets of synaptic weights in place of the internal values we would get if we simply fed $x_n$ into the NN (with current weights). In other words, we have:\\
\medskip

\noindent Layer 1 to 2:
$\ \displaystyle \begin{array}{rcl}
    A^1_n &=&x_n (x_n)^T\\[0.2em]
    b^1_n &=& x_n (\hat z^1_{k_{j_n}})^T
\end{array}$\medskip\\
Layer 2 to 3:
$\ \displaystyle \begin{array}{rcl}
    A^2_n &= & \hat h^1_{k_{j_n}} (\hat h^1_{k_{j_n}})^T\\[0.4em]
    b^2_n &= & \hat{h}^1_{k_{j_n}} (\hat z^2_{k_{u_n}})^T
\end{array}\ \ $.\medskip\\
Layer 3 to 4:
$\ \displaystyle \begin{array}{rcl}
    A^3_n &= & \hat h^2_{k_{u_n}} (\hat h^2_{k_{u_n}})^T\\[0.4em]
    b^3_n &= & \hat h^2_{k_{u_n}} (z^3_n)^T
\end{array}$
\medskip\\

\noindent Otherwise, the algorithm remains essentially the same.

\section{Regression Comparisons}\label{S:Reg}

Given that the updates on the weights are made several samples at a time, our method could be viewed as a counterpart to (batch) gradient descent. However, BLA has the advantage of not curve-following, which appears to help the method avoid getting stuck in `saddle points' and to produce far superior results than GD in regression applications (as we will see from our experiments), so we also included comparisons to online, real-time updating methods like ADAM.  Indeed, BLA appears to require few training data points to achieve a high level of accuracy, making it very suitable for online, real-time updating. We compared our results to three other popular algorithms with respect to speed of convergence: these were gradient descent (GD),  ADAM and the Limited Memory Broyden–Fletcher–Goldfarb–Shanno algorithm (LBFGS). 
\\

We trained these four algorithms (BLA included) on the same data and afterwards calculated the MSE on the same independently generated validation set. The training set for all experiments consists of $\mathcal N=6000$ data points while the validation set was of size $1000$. Each algorithm was tested $1000$ times and the results were averaged to avoid the influence of any outlier.
All $6,000,000$ points were generated by taking each $x$ to be uniformly distributed over the domain of the relevant function $f$ each time, and independent of previously generated $x$'s, and then by setting $y=f(x)$.
For all methods we used the scikit learn MLPRegress0r and MLPClassifier packages in python \cite{Scikit-2011}. For GD we used the ``SGD'' option with batchsize equal to $6000$. For ADAM and LBFGS we used the default settings. 
\\

Our focus herein was on SHLNN with $m=100$ nodes in the single hidden layer. 
We always take $\sigma_2(x)=x$ and $\sigma_1(x)=\tanh(x)$, except for the experiment below on a 
stochastic network, which would be expected to have lots of corner-resembling regions in its graph,
and in the case of which it made sense to work with $\sigma_1(x)=\text{ReLU}(x)$.\\

Our purpose was not to optimize BLA on each problem so as to potentially
give it some advantage.
Rather, we made several rather arbitrary decisions on our BLA implementation. We always took $r=0$ for the
first batch, and otherwise $r=N_m$, so as to weight the previous estimate equally with what would be produced from the new iteration.
We always divided the first epoch into $10$ batches of equal size, the 
second epoch into $5$ batches, the third into $3$, the fourth into $2$, and all remaining epochs
consisted of $1$ batch.
(Experiments and intuition suggest it could have been better to make the batches within an epoch of different sizes, but we have deferred exploring this to future work.)
We also always used the (squared) $\ell_2$ norm for our distance formula (except in the multi-input experiment, where we used the $\ell_\infty$ norm), and took the scoring function to be $s(x)=\exp(-x^2)$ for simplicity.
As already mentioned, for the number $\delta=\delta_m$ of particles that would be considered in the sampling for each actual data point, we started with $\delta_1=40$, and for each subsequent batch we decreased it by 1, until $\delta_m$ became equal to $8$ (afterwards we kept it unchanged).

\medskip

\begin{srem}
 For functions where the input $x$ or the output $y$ (or some component of the input $x$) are orders of magnitude larger than the other value(s), it is advised that we first scale to the same range of values so that the distance between corresponding inputs or outputs is not so much greater than the other distance(s) in terms of magnitude.  None of our functions required this.\\
 \end{srem}
\medskip

\subsection{Single Input, Single Output Regression}

For these experiments, we consider SHLNN regression for a continuous function $f:K\to \mathbf{R}$ that takes a single input from a compact subset $K$ and outputs a single real number.
In particular, we compare the four algorithms first on the following three
functions:
\begin{align}\label{E:single}
    f_1(x) &= x^3 - 2x^2 +5x - 1, \text{ where } x \in {[-3,3]}\\
    f_2(x) &=\sin(x^2)-0.03x^5, \text{ where } x \in {[-3,3]}\\
    f_3(x) &=-(x-2)^3(x+1)^2(x-4)/8, \text{ where } x \in {[-1,4]}.
\end{align}

\noindent The following graph shows the MSE of one trial over the training period:\\

\begin{figure}[H]
    \centering
    \includegraphics[scale=0.45]{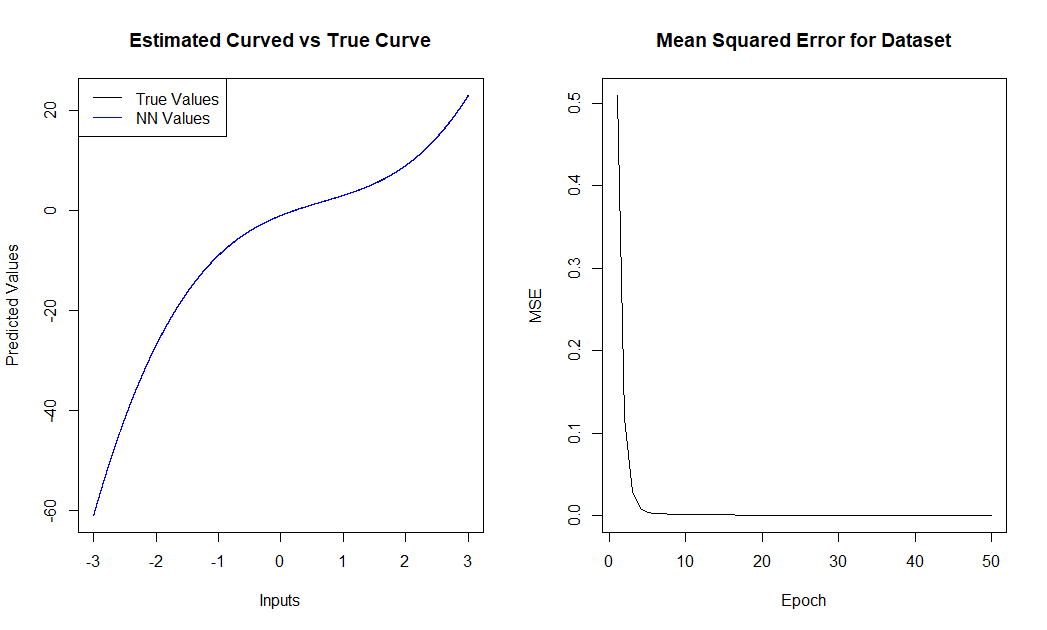}
    \caption{The Approximation and MSE of $x^3 - 2x^2 +5x - 1$ up to $50$ epochs of the training set.}
    \label{fig:polytraining}
\end{figure}

The average MSE comparison up to $50$ epochs for the four algorithms and the functions $f_1$, $f_2$ and $f_3$ can be found in the following three tables respectively:

\begin{center}
\begin{tabular}{ c | c | c |  c | c }
\Xhline{3\arrayrulewidth}
{} & \multicolumn{4}{c}{$x^3 - 2x^2 +5x - 1$, where $x \in {[-3,3]}$} \\
\cline{2-5}
Epoch  & BLA & GD  & ADAM & LBFGS \\ 
\hline
 1 & $0.1819$ & $406.372$  & $374.767$ &  $314.949$ \\  
 5 & $0.0099$ & $359.012$  & $202.644$ & $73.127$ \\
 10 & $0.0065$ & $303.739$ & $75.947$ & $53.362$ \\
 15 & $0.0050$ & $251.660$ & $58.347$ & $45.512$ \\
 25 & $0.0035$ & $166.631$ & $40.492$ & $30.194$ \\
 50 & $0.0019$ & $87.064$ & $15.262$ & $11.001$ \\
 \Xhline{3\arrayrulewidth}
\end{tabular}
\end{center}

\begin{center}
\begin{tabular}{ c | c | c | c | c }
\Xhline{3\arrayrulewidth}
{} & \multicolumn{4}{c}{$\sin(x^2)-0.03x^5$, where $x \in {[-3,3]}$} \\
\cline{2-5}
Epoch  & BLA & GD  & ADAM & LBFGS \\ 
\hline
 1 & $0.1104$ & $5.333$ & $2.686$ & $3.049$ \\  
 5 & $0.0886$ & $4.884$ & $2.022$ & $2.036$ \\
 10 & $0.0869$ & $4.410$ & $1.946$ & $2.015$ \\
 15 & $0.0857$ & $4.016$ & $1.800$ & $1.987$ \\
 25 & $0.0838$ & $3.420$ & $1.198$ & $1.642$ \\
 50 & $0.0805$ & $2.616$ & $0.785$ & $1.053$ \\
 \Xhline{3\arrayrulewidth}
\end{tabular}
\end{center}

\begin{center}
\begin{tabular}{ c | c | c | c | c }
\Xhline{3\arrayrulewidth}
{} & \multicolumn{4}{c}{$-(x-2)^3(x+1)^2(x-4)/8$,}\\ 
{} & \multicolumn{4}{c}{where $x \in {[-1,4]}$} \\
\cline{2-5}
Epoch  & BLA & GD & ADAM & LBFGS \\ 
\hline
 1 & $0.060$ & $5.997$  & $3.822$ & $4.271$ \\  
 5 & $0.038$ & $5.740$ & $1.670$ & $1.938$ \\
 10 & $0.035$ & $5.468$ & $1.493$ & $1.696$ \\
 15 & $0.033$ & $5.240$ & $1.241$ & $1.607$ \\
 25 & $0.028$ & $4.875$ & $0.832$ & $1.157$ \\
 50 & $0.022$ & $4.247$ & $0.467$ & $0.654$ \\
 \Xhline{3\arrayrulewidth}
\end{tabular}
\end{center}
In the last two cases it is believed that BLA may be approaching the optimal
lower bound (for our $100$-hidden-states NN), causing its improvement to slow down. 

\subsection{Random Weight Neural Network}

The previous examples dealt with smooth functions, which might make them quicker to approximate. To deal with a considerably more complicated function, we generated a data set using a function that was defined via a neural network itself. In particular, we constructed a separate single-input, single-output neural network (which we'll denote by ${\rm NN}_0$ here) with the same architecture as for our previous examples (that is, having one hidden layer with $m=100$ nodes). The weights and biases of the first layer were assigned through a normal random variable with mean $5$ and variance $3$. Similarly, the second layer was assigned weights and biases normally with mean 0 and variance $0.5$. All assignments were independent of others.

\smallskip

For the first activation function of ${\rm NN}_0$, we chose the hyperbolic tangent activation function, while the second activation function was the identity. After we set up ${\rm NN}_0$, we uniformly generated inputs from ${[-5,5]}$ and passed them through the network. The resulting pairs of input and output values formed the training and validation sets that we used to run our experiment.\\

Now, regarding the network ${\rm NN}_1$ that we want to train to approximate ${\rm NN}_0$ (as a function), we have to remark the following: since a function defined as above, through a neural network such as ${\rm NN}_0$, will have a graph that does not look smooth at all, it appears to be generally better (for all methods tested here) to choose ${\rm NN}_1$ to have an activation function that also has a corner.
Hence, instead of the first activation function being $\tanh$ (as in our previous experiments, and somewhat oddly also as in the definition of ${\rm NN}_0$ itself), we used ReLU (the second activation function of ${\rm NN}_1$ was again the identity). 

\smallskip

An example training set can be seen in the figure below. 

\begin{figure}[H]
    \centering
    \includegraphics[scale=0.45]{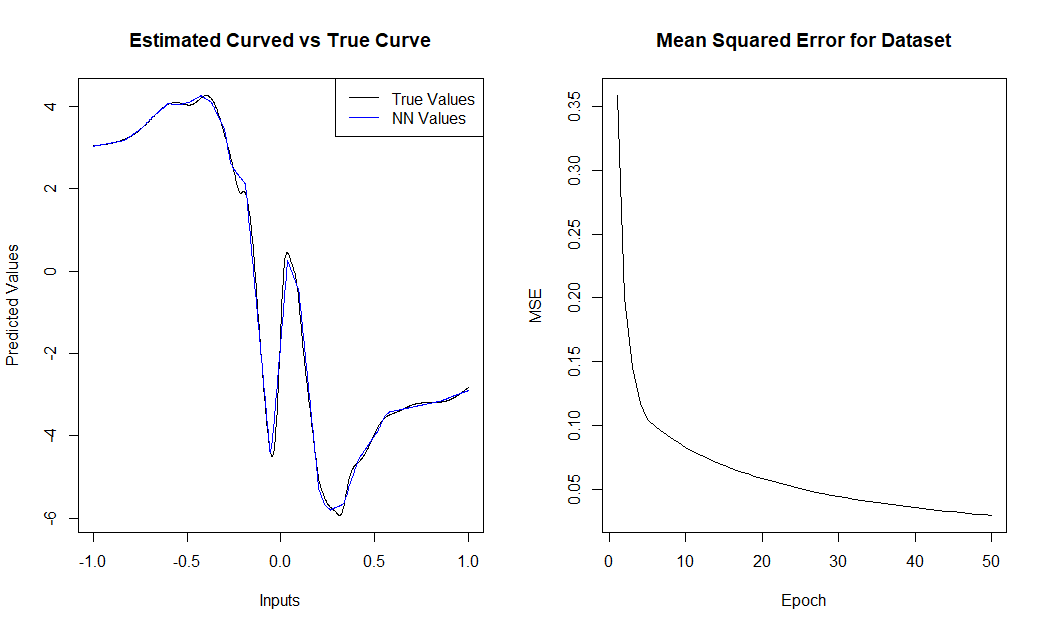}
    \caption{Example run of BLA on stochastically generated neural network over $50$ epochs. }
    \label{fig:Tan_network}
\end{figure}

The MSE comparison was as follows:

\begin{center}
\begin{tabular}{ c | c | c | c | c }
\Xhline{3\arrayrulewidth}
{} & \multicolumn{4}{c}{Stochastic Network} \\
\cline{2-5}
Epoch  & BLA & GD & ADAM & LBFGS \\ 
\hline
 1 & $0.203$ & $39.749$  & $29.645$ & $16.802$ \\  
 5 & $0.088$ & $31.891$ & $6.749$ & $4.648$ \\
 10 & $0.073$ & $24.894$ & $2.516$ & $3.176$ \\
 15 & $0.066$ & $20.051$ & $1.304$ & $2.214$ \\
 25 & $0.056$ & $14.402$ & $0.722$ & $1.269$ \\
 50 & $0.045$ & $9.518$ & $0.306$ & $0.628$ \\
 \Xhline{3\arrayrulewidth}
\end{tabular}
\end{center}

Similarly to the previous experiments, the BLA can quickly find good estimates for this stochastically generated function. Again, it is speculated that the appearance of a slowdown in the rate of the convergence of BLA would be due to its approaching the theoretical lower bound.

\subsection{Multiple Inputs, Single Output}

When increasing the dimension of the problem, we should expect more data to be required in order to get reasonable estimates. We should also expect that we would have to make more adaptive choices when it comes to (hyper)parameters such as $r=N_m$ and $\delta_1=40$.
However, we wanted to keep one universal set of parameters throughout this paper to show that BLA is a robust method.

\smallskip

The function we approximate here is:
\begin{align*}
    f(x_1,x_2,x_3) = 2{x_1}^2 x_2-6x_1x_3\text{ where } x_1 \in {[-5,5]},\, x_2 \in {[-2,2]},\, x_3 \in {[0,4]}
\end{align*}
and the BLA MSE starts at about $200$ after one epoch and quickly converges from there:

\begin{figure}[H]
    \centering
    \includegraphics[scale=0.4]{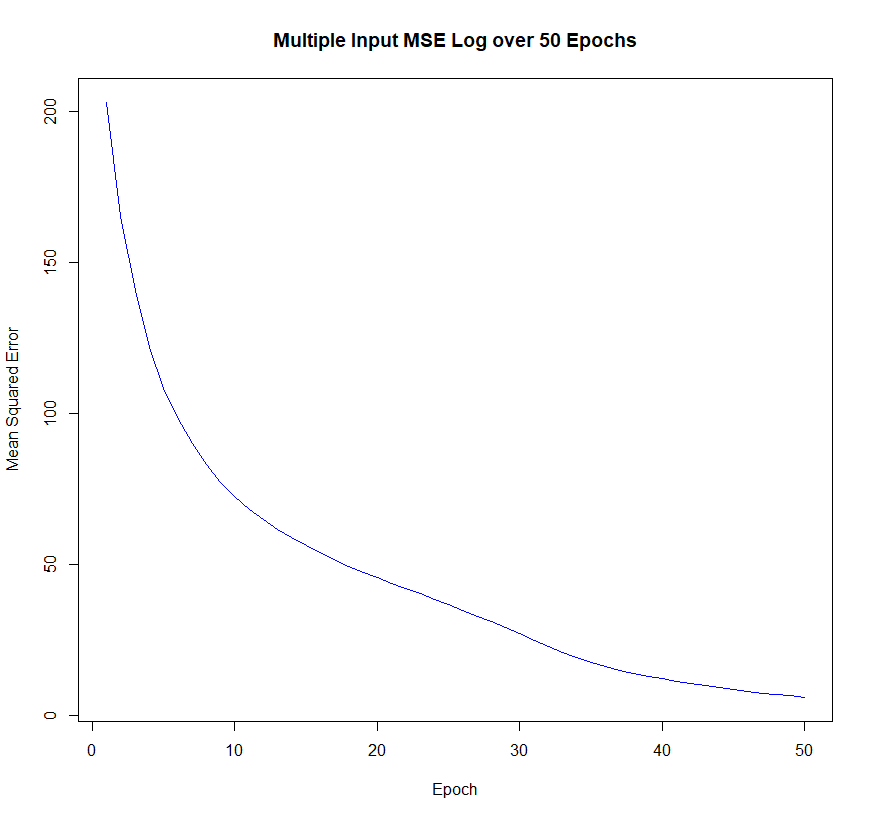}
    \caption{Mean squared error for an example run of BLA on three-dimensional input data set. }
    \label{fig:Tan_network}
\end{figure}
The comparison with the other methods is as follows:
\begin{center}
\begin{tabular}{ c | c | c | c | c }
\Xhline{3\arrayrulewidth}
{} & \multicolumn{4}{c}{Multi-Input} \\
\cline{2-5}
Epoch  & BLA & GD & ADAM & LBFGS \\ 
\hline
 1 & $203.044$ & $2185.198$  & $2135.537$ & $1987.212$ \\  
 5 & $108.205$ & $1860.787$ & $1641.746$ & $873.113$ \\
 10 & $72.536$ & $1492.288$ & $1153.364$ & $732.618$ \\
 15 & $56.370$ & $1247.324$ & $922.918$ & $648.826$\\
 25 & $36.803$ & $1053.292$ & $671.781$ & $529.465$ \\
 50 & $6.123$ & $840.075$ & $463.275$ & $390.943$ \\
 \Xhline{3\arrayrulewidth}
\end{tabular}
\end{center}

Clearly, BLA did not `finish' early here.  Its outperforming the other methods seemed to accelerate, which may be due to the size of the multiple-input problem. 

\section{Binary Classification}\label{S:Class}

We have claimed that BLA is a supervised learning method but hitherto we have only considered a class of regression problems.
Hence, we now want to demonstrate BLA working on another type of problems, and in particular on binary classification. Typically, these problems are solved by a method like logistic regression, which fits a (sigmoid) curve assigning probabilities to each point that predict which class an observation belongs to. Of course it is known that we could also train a neural network to fit such a curve, although we should expect the resulting curve to now have a more complicated graph. Thus we can test BLA on the examples below. 

\medskip

As always, we use data sets that consist of a training set of $6000$ observations and a validation set of size $1000$. 
We first consider a simple data-set of one predictor for two defined classes. The data for these classes were generated through four independent Bernoulli distributions that are defined on four different intervals. Specifically, we generate a random value from the entire domain (containing all four intervals), and based off its value we assign it to one of the four Bernoulli distributions to classify it as a $0$ or $1$. This allows us to generate well-separated data which also contain some noise. 

For this example, we consider our predictor to be defined on ${[0,1]}$. For the two classes generated, we want values closer to $x=0$ to be assigned the class $y=0$ with a high probability, whereas values closer to $x=1$ should more often be mapped to $y=1$. Thus, we used the following function to generate our data set:
\begin{align*}
\begin{array}{ll}
      \text{Bernoulli}(0.05), & \text{for } x< 0.3 \\
      \text{Bernoulli}(0.25), & \text{for } 0.3 \leq x< 0.6 \\
      \text{Bernoulli}(0.75), & \text{for } 0.6\leq x< 0.8 \\
      \text{Bernoulli}(0.95), & \text{for } 0.8\leq x \\
\end{array} .
\end{align*}

For example, if we generate a random sample to be $x_1 = 0.5$, this sample's corresponding $y$ value will be the result of a Bernoulli$(0.25)$. After $300$ of the $6000$ data points in the training set, the data looks like this:
\begin{figure}[H]
    \centering
    \includegraphics[scale=0.35]{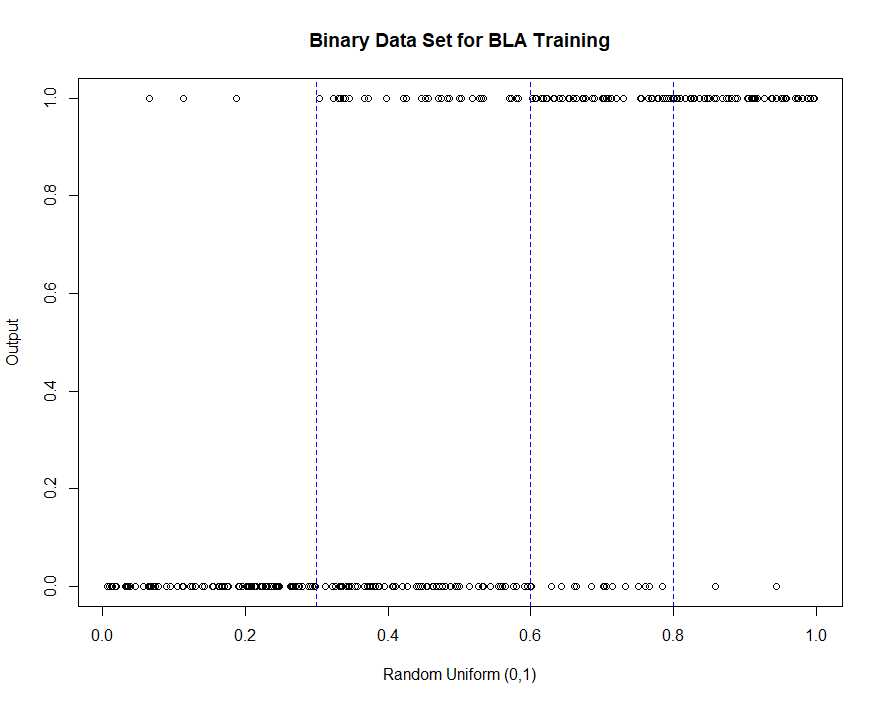}
    \caption{Example training set of $300$ units generated from our probability distribution.}
    \label{fig:BinaryClassifer}
\end{figure}

To predict the classes, we first use this data and BLA to train a SHLNN with
all our standard settings as above.
This SHLNN function $g(x)$ will produce real numbers concentrating in $[0,1]$ for
$x\in [0,1]$.
From there we pick a cut-off value $c=0.5$, so that any $g(x)\ge 0.5$ is predicted to be class $1$ and otherwise class $0$.\\

To demonstrate that BLA convergence essentially takes place within one epoch, we ran $1000$ trials of $1$ epoch and with this cut-off value. On average BLA achieved an accuracy of $84.364\%$.\\  

Given that in this situation we know the underlying formula that generated this data set we can calculate an upper limit for the accuracy (which comes from the fact that there is always some `noise', or in other words stochasticity). If we were to assign a value of $1$ for $x\geq 0.6$ and $0$ otherwise, and simulate the data thousands of times, we can thus calculate that our upper limit to the prediction accuracy would be $85\%$. So after only $1$ epoch of the data, BLA is below this upper limit by only $0.636\%$.\\

For a second data set we wanted a binary data set where multiple (disjoint) sub-regions of the domain can be assigned the class $1$ with high probability. To create this, we pass values from a ${[0, 2\pi ]}$-uniform distribution into a specific ${[0,1]}$-valued function which then component-wise specifies a different Bernoulli distribution.  Hence, our output is a family of Bernoulli random variables with probability $p(x) = (\cos(x)+1)/2$ for $x \in {[0, 2\pi ]}$. This function leads to having two end sub-regions of the domain, points from which will be classified as a $1$ with high probability, and a middle sub-region which is classified as a $0$ with high probability. Here is an example of a training set with $300$ observations:

\begin{figure}[H]
    \centering
    \includegraphics[scale=0.35]{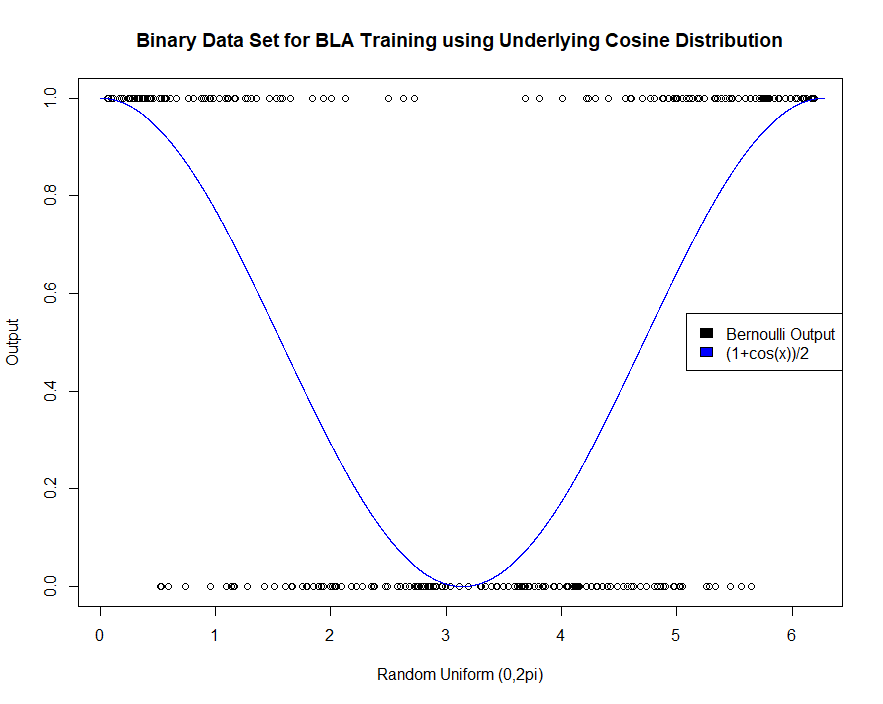}
    \caption{\small Example training set of $300$ units generated using the (1+cos(x))/2 probability distribution.}
    \label{fig:BinaryClassifer}
\end{figure}
Similarly to before, we simulate an upper limit to the accuracy for this data set. To do this we generated thousands of data sets, and if any unit had a probability greater than $0.5$ given our probability function, we assigned it the value $1$, otherwise the value $0$. With a cut-off value of $c=0.5$ we estimated that we could achieve an accuracy of $81.83\%$ on average. After running our method for only one epoch for $100$ different initializations, we achieve an accuracy of $81.114\%$.{ Compare this to ADAM, where, if we run this data set $100$ times, we only achieve an accuracy of $60\%$ after one epoch (the accuracy does become comparable later, at least with ADAM, reaching $81.6\%$ on average in our experiment after 100 epochs). A full breakdown of the four methods can be seen below (we worked with the hyperbolic tangent as our first activation function for each).}\\

\begin{center}
\begin{tabular}{ c | c | c | c | c }
\Xhline{3\arrayrulewidth}
{} & \multicolumn{4}{c}{2nd Classification Example} \\
\cline{2-5}
Epoch  & BLA & GD & ADAM & LBFGS \\ 
\hline
 1 & $81.114\%$ & $50.284\%$  & $60.045\%$ & $56.842\%$ \\  
 5 & $80.98\%$ & $50.590\%$ & $78.575\%$ & $64.053\%$ \\
 10 & $81.122\%$ & $51.585\%$ & $80.874\%$ & $70.595\%$ \\
 15 & $81.387\%$ & $51.985\%$ & $81.212\%$ & $70.449\&$ \\
 25 & $81.584\%$ & $55.766\%$ & $81.461\%$ & $75.591\%$ \\
 50 & $81.763\%$ & $58.106\%$ & $81.583\%$ & $79.966\%$ \\
 \Xhline{3\arrayrulewidth}
\end{tabular}
\end{center}

\section{Conclusions and Future Work}

In this paper, we introduced the Bootstrap Learning algorithm (BLA), which we propose for supervised learning and, in particular, for training wide neural networks. We have 
tested the method in some regression and binary classification problems. 

For a wide neural network, this bootstrap learning method can deliver quick and accurate estimates upwards of $100$ times more precise (in terms of MSE) than other common machine learning techniques after a similar number of epochs.  

The method needs to be investigated further from a point of view of applicability, convergence and efficiency of use.
For instance, we need to adapt its current implementation in order to apply and test it on other more challenging problems, such as learning LSTMs. We should also investigate its convergence (and speed of convergence) when the method is applied to deep neural networks, which should also allow for a multitude of other applications. 

There is expectation that an appropriate online version of the current (batchmode) implementation can be developed, This should also be of great use, as suggested by the already shown quick convergence of the method which can be achieved with only few training data. 

Finally, the mathematical foundations of BLA need to be established.


\end{document}